\documentclass[acmtog,sigconf]{acmart}
\usepackage{float}
\usepackage{comment}
\AtBeginDocument{%
  \providecommand\BibTeX{{%
    \normalfont B\kern-0.5em{\scshape i\kern-0.25em b}\kern-0.8em\TeX}}}

\setcopyright{acmcopyright}
\copyrightyear{2022}
\acmYear{2022}
\acmDOI{XXXXXXX.XXXXXXX}

\acmConference[ICONS '22]{International Conference on Neuromorphic Systems}{July 27--29,
  2022}{Knoxville, Tennessee}
\acmBooktitle{ICONS '22: International Conference on Neuromorphic Systems,
 July 27--29, 2022, KNOXVILLE, TENNESSEE} 

\begin{document}

\title{Learning over time using a neuromorphic adaptive control algorithm for robotic arms}

\author{Lazar Supic}
\email{lazar.supic@accenture.com}
\affiliation{%
  \institution{Accenture Labs}
  \streetaddress{415 Mission Street}
  \city{San Francisco}
  \state{California}
  \country{USA}
  \postcode{94105}
}

\author{Terrence C. Stewart}
\email{terrence.stewart@nrc-cnrc.gc.ca}
\affiliation{%
  \institution{National Research Council Canada}
  \streetaddress{1200 Montreal Road, Building M-50}
  \city{Ottawa}
  \country{Canada}}








\begin{abstract}
  In this paper, we explore the ability of a robot arm to learn the underlying operation space defined by the positions (x, y, z) that the arm’s end-effector can reach, including disturbances, by deploying and thoroughly evaluating a Spiking Neural Network SNN-based adaptive control algorithm. While traditional control algorithms for robotics have limitations in both adapting to new and dynamic environments, we show that the robot arm can learn the operational space and complete tasks faster over time. We also demonstrate that the adaptive robot control algorithm based on SNNs enables a fast response while maintaining energy efficiency. We obtained these results by performing an extensive search of the adaptive algorithm parameter space, and evaluating algorithm performance for different SNN network sizes, learning rates, dynamic robot arm trajectories, and response times. We show that the robot arm learns to complete tasks 15\% faster in specific experiment scenarios such as scenarios with six or nine random target points. 
\end{abstract}




\maketitle

\section{Introduction}

Robotic arms are becoming increasingly prevalent in applications central to human life, including manufacturing, rehabilitation, and a growing range of household tasks as assistive devices\cite{erol2007smooth,naotunna2015meal,xiao2012smooth}. Despite the fact that the capabilities of robotic arms have expanded quickly in recent years and have reached very strong performance in repetitive, prescribed tasks, their ability to handle unexpected situations, be flexible, and adapt is still quite poor compared to biological organisms. Therefore, one of the main questions in robotics is how we can enable robotic arms to learn and execute flexibly and adapt to new and dynamic environments, while preserving the energy efficiency of robotic arm execution of fixed tasks.

\begin{figure*}[!h]
  \includegraphics[width=\linewidth]{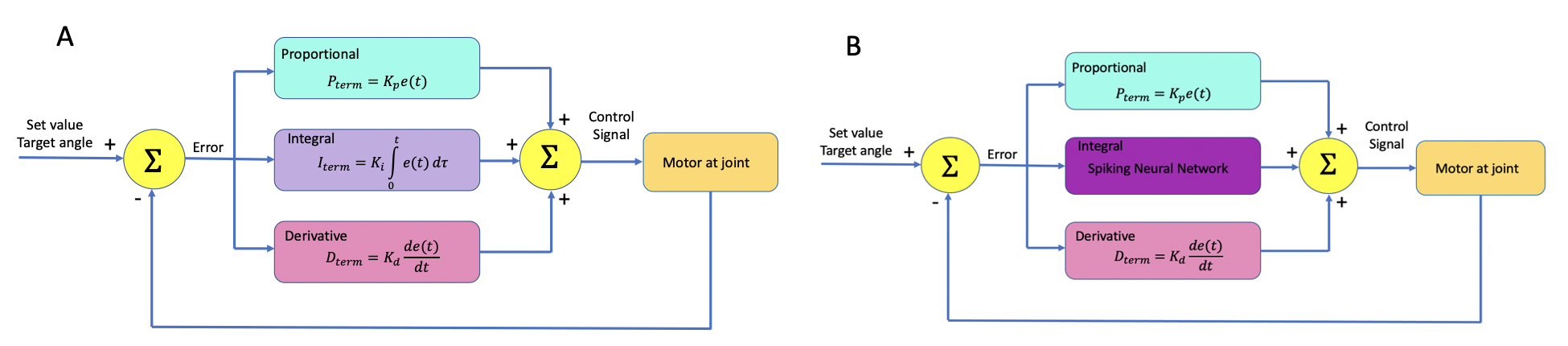}
  \caption{\textbf{A.} Canonical PID block diagram. \textbf{B.}Spiking Neural Network (SNN) based diagram }
  \Description{Results for different experiments.}
  \label{fig:results_neuronal_ensamble_size}
\end{figure*}

Neurorobotics is an emerging branch in robotics that combines neuroscience, robotics, and artificial intelligence with the key goal of embedding brain-inspired algorithms into robots to enable them to learn better and handle these more complex, dynamic situations. The most prominent among these brain-inspired algorithms are Spiking Neural Networks (SNNs), which are artificial neural networks modeled after principles of biological spike-based brain processing. It has been shown that adaptive robot arm controllers implemented using SNNs improve spatial accuracy and energy efficiency for typical robot arm tasks, such as reaching, compared to canonical control algorithms ~\cite{DeWolf2016adaptive,zaidel2021neuromorphic}. However, open questions remain about the underlying learning mechanisms of SNNs, including their ability to learn an operation space, their learning rates over time, as well as the roles of pretraining and online learning for handling disturbances in the operation space, such as a change in the trajectory or a change in the weight of the object being handled by the robotic arm.

A robot arm operates in a three-dimensional operational control space defined by positions (x, y, z) that the arm’s end-effector can reach. The ultimate goal of control algorithms is to bring the robot arm end-effector (Fig. 1) to the desired target position in this space. To achieve this, proper control commands must be given to the motors at the robot arm joints. An inverse kinetic algorithm can compute the angle values at the robot arm joints to reach a given target location. In theory, these values would be enough to execute the reach task flawlessly. However, due to real-world limitations - including motors at the joints having their own dynamics (defined by their transfer functions), friction at the joints, as well as the weight of the whole system dynamically changing when the arm starts carrying a heavy object – the ability of the robot arm to reach the target position accurately is compromised.

In this paper, we explore the ability of a robot arm to learn the underlying operation space, including disturbances, by deploying and thoroughly evaluating an SNN-based adaptive control algorithm ~\cite{DeWolf2016adaptive}. We find that the neuromorphic adaptive control algorithm can learn over time and move between the target destination points faster as it learns. We varied several parameters that affect algorithm performance to obtain these results, including SNN size, learning rate, and reaching task scenarios across the operation space. The key quantitive result achieved in the paper is that we showed that the robot arm learns to complete
tasks 15\% faster in specific experimental scenarios.  While the underlying algorithm is one that has been previously presented (\cite{stewart2015closed}~\cite{dewolf2017} , our contribution here is to vary these parameters in order to discover how the system performs across that space.



\section{Related Work}



\subsection{Classical PID}

A traditional way to deal with unknown aspects of a motor control problem has been to use a Proportional–Integral–Derivative (PID) controller, shown in Fig. 2. An error signal is first computed as the difference between the target and current angle of the robot arm joint.  The PID controller then processes this error signal in three parallel branches: in the proportional branch, the error signal is multiplied by the $K_{p}$ proportional constant; in the integral branch, the error signal is integrated and multiplied by the constant $K_{i}$, and the derivative branch takes the first derivative of the error signal, multiplied by the constant $K_{d}$~\cite{1453566}. The constants $K_{p}$, $K_{i}$, and $K_{d}$ are fixed in advance based on the dynamics of the system and expected disturbances. For a well-defined, static environment, this control algorithm would perform reasonably well. However, in dynamic environments, this type of control would not be able to adapt to changing conditions. Instead, it is common to manually re-tune these parameters across different conditions.




\subsection{SSN-based adaptive control}

To enable robot arms to learn, we used a published adaptive PID algorithm where the integral branch of the PID is implemented as a Spiking Neural Network(SNN) using the dynamics adaptation approach \cite{DeWolf2016adaptive}. We note that we selected this algorithm because it has the potential to learn over time, which is not a common approach in the past. Other algorithms have fixed architecture and predefined/tuned parameters. In the SNN based approach the transfer function between the error signal and the integral part of the control signal is governed by the SNN. SNN connection weights are learned in real-time using the Prescribed Error Sensitivity (PES) rule\cite{Voelker2015pes}. The Neural Engineering Framework (NEF)~\cite{10.3389/fninf.2013.00048} and Nengo are used as a framework to implement the SNN. Two parameters of the SNN are of particular interest to this study: the number of neurons used to implement the SNN and the learning rate for the PES rule. The proportional coefficient for the adaptive controller $K_{p}$ is 200, and for the derivative branch is multiplied by coefficient $K_{d}$ = 10. The coefficients were selected via iterative tuning, in accordance with typical PID controller coefficient selection. The adaptive controller is a
patented algorithm from ABR and is available at the following link~\cite{abr_control}. 




\section{Methods}
We evaluated the online learning capability of the adaptive control on the Jaco 6 DOF robotic arm. The key question we want to answer is whether and how the loaded arm moving between two random points in the operational space can 'learn' to execute this task faster. We utilized the MuJoCo robotic simulator to execute simulations of the physical characteristics of the robot arm, including movement dynamics. Specifically, the MuJoCo simulator contains a physics-based model of the Jaco robotic arm. The dynamic and physical characteristics of the Jaco arm are described in ~\cite{6386109}. We impose a firm time limit that tasks should be completed within 2000 simulation steps, equivalent to 2s in the real-world scenario. This time limit was selected to give the robot arm an upper bound on timing to complete the task. If the task is not completed, the arm gets a new task, i.e. a new target.

We divided the simulation cycle into two phases: the training phase and the testing/evaluation phase. In the training phase, we ran 20 simulations per transition between two points. We focused on varying the learning rate parameter because it is primarily responsible for the learning capability of the algorithm. Similarly, because the focus of the study was the learning capability, the neuron ensemble size was varied. The simulation task was the reaching task, which is common in robotics applications. The trajectories between targets were selected to be random, in order to sample the operational space. Specifically, the randomly chosen points were enumerated, and the arm moves between them in increasing numerical order. For example, if we have five randomly chosen target points, they are enumerated as 1,2,3,4 and 5. During the training phase, we move the robot arm's end-effector in the following order 1->2->3->4->5->1. The robot arm was loaded with an equivalent weight of 1kg. The time to reach the target point was measured and plotted. We randomly picked one of the targets from the training phase during the testing phase and measured the time the arm's end-effector needed to reach the target and plotted the histogram of the resulting times. 

In the evaluation phase, we varied several parameters: the size of the neuron ensemble, the PES learning rate, and the number of target positions to determine their impact on learning over time. We found that for our SNN-based robot arm, the control algorithm has optimal performance with 5000 neurons and a learning rate 5e-5.

\section{Results}
\subsection{Experiment with different size of neuron neuron ensemble}

We started with a simple case of three random target points and used the adaptive PID controller described in the Methods section to complete the reach task. We tried different sizes of the neuronal ensemble (i.e., different sizes of the Spiking Neural Network): 2500, 5000, and 10000 neurons. The results are shown in the Fig ~\ref{fig:results_neuronal_ensamble_size}, where the training phase results are shown in blue histograms, and the results from the testing phase are shown in red histograms. The graphs show how different ensemble sizes affect the learning capability of the algorithm. Based on the timing results and simulation duration, we picked 5000 neurons as the optimal SNN size based on two criteria: 1) simulation time and 2) the number of tasks completed during this time. Specifically, it takes 20 cycles to do the training phase for nine targets, which is equivalent to 180 tasks. The testing phase includes 1000 tasks. For the case of 10000 neurons, this whole simulation took 12 hours. Another essential criterion from the implementation efficiency perspective is to reduce the number of neurons. We thus tested two lower numbers -  2500 neurons, and 5000 neurons. The simulation with 2500 was slightly faster, but the number of completed tasks was 10-15\% lower than with 5000 neurons. This is why we decided to run the rest of the simulations with 5000 neurons.

\begin{figure*}
  \centering
  \includegraphics[width=\linewidth]{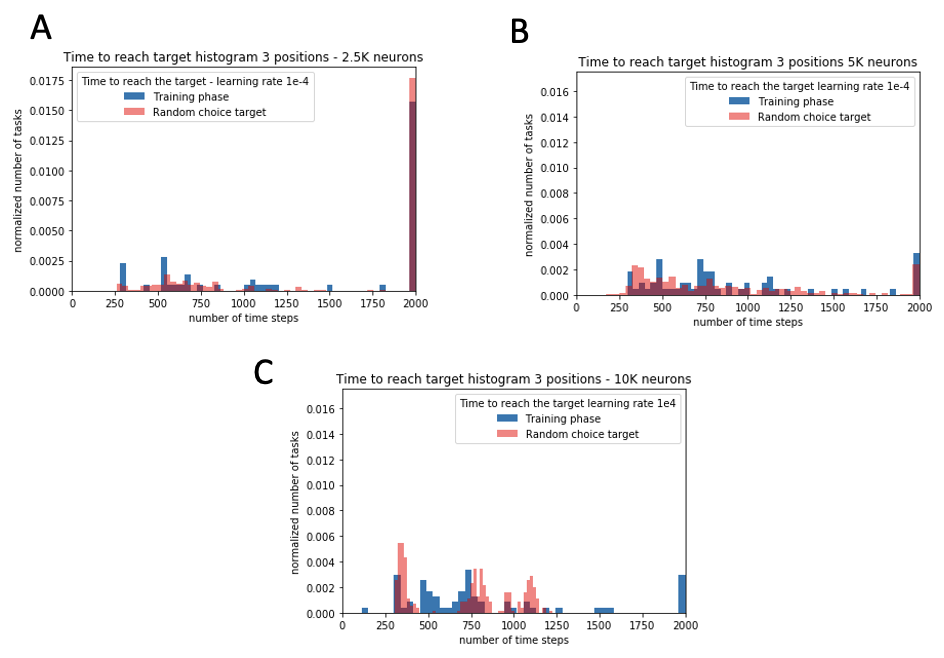}
  \caption{Time to reach the target diagrams with different size of neuron ensemble \textbf{A.} The size of neuronal ensemble of 2500 neurons. \textbf{B.} The size of neuronal ensemble of 5000 neurons. \textbf{C.} The size of neuronal ensemble of 10000 neurons. }
  \Description{Results for different experiments.}
  \label{fig:results_neuronal_ensamble_size}
\end{figure*}


\subsection{Experiment with different PES learning rate}

\begin{figure*}[h!]
  \centering
  \includegraphics[width=\linewidth]{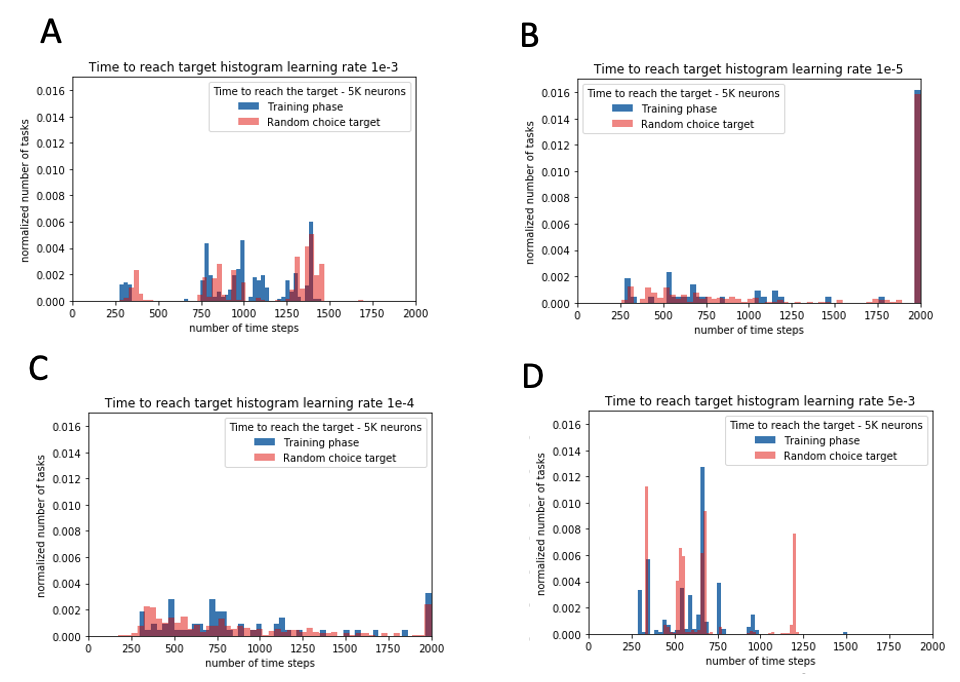}
  \caption{Time to reach the target diagrams with different learning rate \textbf{A.} The PES learning rate of e-3. \textbf{B.} The PES learning rate of e-5.  \textbf{C.} The PES learning rate of e-4.  \textbf{D.} The PES learning rate of 5e-5 }
  \Description{Results for different experiments.}
  \label{fig:results_learning_rate}
\end{figure*}

To explore how the learning capability of the adaptive algorithm depends on the PES learning rate, we tried four PES learning rates: 1e-3, 1e-4, 5e-5, 1e-5. The results are shown in the Fig ~\ref{fig:results_learning_rate}. As shown Fig ~\ref{fig:results_learning_rate} A, with a learning rate of 1e-3, around 1400 simulation steps are needed to complete the task, which is quite long. Moreover, as shown by the strong peak in Fig ~\ref{fig:results_learning_rate} B, the task was not completed with a learning rate of 1e-5. An optimized learning rate of 5e-5 was chosen, which enabled the simulation to complete in under 1000 steps, as shown in Fig ~\ref{fig:results_learning_rate} D.

\subsection{Experiment with different number of target positions}

\begin{figure*}[h]
  \centering
  \includegraphics[width=\linewidth]{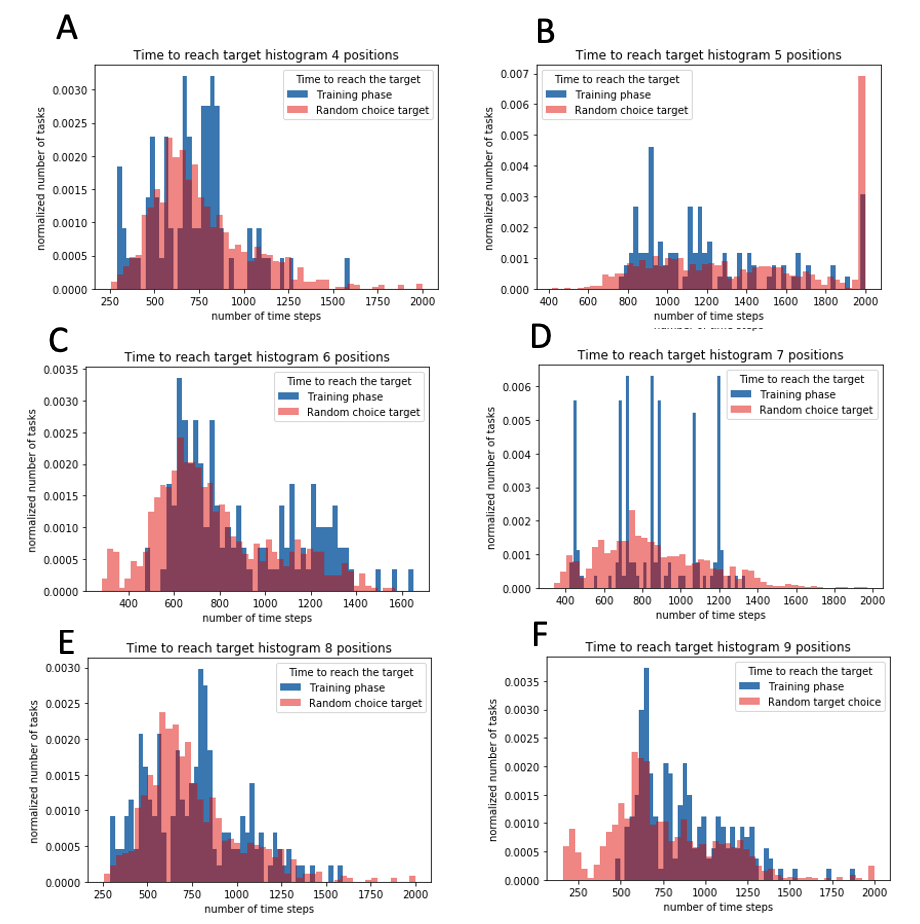}
  \caption{Time to reach the target diagrams with number of target points using the adaptive PID with 5000 neurons and a PES learning rate of 5e-5. \textbf{A.} Four random target points. \textbf{B.} Five random target points.  \textbf{C.} Six random target points.  \textbf{D.} Seven random target points  
  \textbf{F.} Nine random target points.}
  \Description{Results for different experiments.}
  \label{fig:results_target_points}
\end{figure*}

Based on our other results, we used an adaptive PID control with 5000 neurons and a PES learning rate of 5e-5 to complete the reach tasks while we increased the number of target points in the simulation. We increased the complexity to nine target points, starting with four random target points. The results of the T-test and bootstrapping test are shown in Fig ~\ref{fig:results_target_points} Table.~\ref{tab:freq}. As shown by the Table, for scenarios with 5, 6 and 9 targets, the robot arm learns to complete tasks 15\% faster. The key take away from this result is that the ability to accelerate learning depends on the target selection strategy, as revealed by the fact that some target selection strategies resulted in accelerated learning over time, while others did not. 

\begin{table*}
  \caption{95\% Bootstrap Confidence Intervals and T-test results for mean time to reach the target }
  \label{tab:freq}
  \begin{tabular}{cccccl}
    \toprule
   Number of targets& Mean Evaluation&Mean training&Bootstrap evaluation interval&Boot strap training interval&T-test p value\\
   \midrule
    4 &757.324& 714.071 &743.93 - 770.68&671.5 - 756.64& >0.05\\
    5 & 1207.44 & 1390.44 &1152,44 - 1262.44&1367.86 - 1413.02&<0.05\\
    6 &777.754 & 893.96&764.12 - 791.15&854.61 - 853.07 &<0.05\\
    7 &853.756 & 850.918 & 839.08 - 868.74 &818.59 - 883.75&>0.05\\
    8 &753.157 & 760.505&740.15 - 767.74&726.54 - 794.69& >0.05\\
    9 &749.965 &
    870.433&732.67 - 767.19&839.79 - 901.78& <0.05\\
  \bottomrule
\end{tabular}
\end{table*}

\subsection{Training vs no-training and learning vs non-learning experiments}

\begin{figure*}[h]
  \centering
  \includegraphics[width=\linewidth]{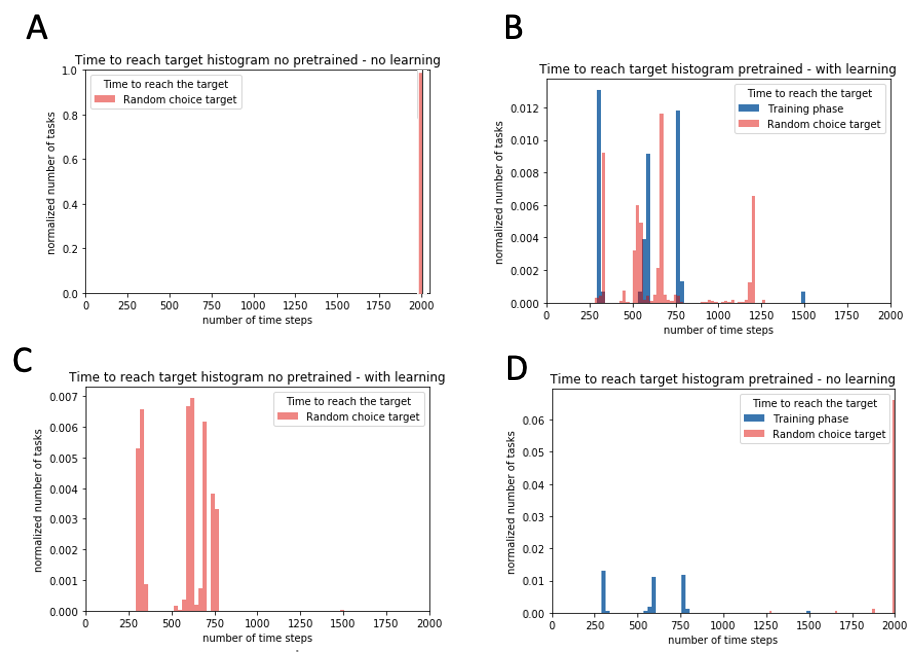}
  \caption{Time to reach the target diagrams training vs no-training. \textbf{A.} PES learning rate is 0. \textbf{B.} PES learning rate is  5e-5 during the training phase, and is 0 during the evaluation phase. \textbf{C.} No training phase, just evaluation phase. \textbf{A.} Training phase and then evaluation phase.}
  \Description{Results for different experiments.}
  \label{fig:results_training}
\end{figure*}

We also evaluated the corner case of what happens if the PES learning rate is zero after the training phase has completed.  That is, during testing we turn off the adaptive term, but still take the output from the network, and add it to the output of the PD controller. (We note that for the special case of learning rate = 0, the integral branch of the PID controller goes to zero, such that the PID effectively becomes a PD controller.)  The goal here is to evaluate whether pre-training might be sufficient for performance, so that the learning rule itself does not have to be used as part of a deployed system. As we can see from Figure ~\ref{fig:results_training} A and B, this does not succeed. All results are clustered at 2000, which means that no task was completed within 2 seconds. This is somewhat expected, as without the adaptation in the neural network, the overall system is merely a PD controller with a state-dependent constant offset applied.  If that state-dependent constant offset is not a \textit{perfect} model of the physics of the environment, then a PD controller is not, in general, going to converge to a particular target location. We also tested the option if we do not do the training phase, i.e. not looping through the target points before starting the evaluation phase (Figures ~\ref{fig:results_training} C and D). The bootstrap test results for the training and evaluation phase experiment are the number of simulation steps to reach the target of 560.63 and 650.85, respectively. The p-value of the T-test is 0.011. In the case where random target points were chosen from the beginning, i.e. there was no pre-training, the result of the bootstrap test is 559.529 simulation steps.

The key insight from the bootstrap analysis is that the training and evaluation phase distributions are clearly separated, as confirmed via both differences in mean values and 95\% confidence intervals. The clear distribution separation in turn means that the algorithm runs faster in the evaluation phase than the training phase. In other words, the bootstrap analysis results reveal the accelerated learning over time for the 5, 6, and 9 target cases shown in Table 1.

\section{Conclusion}

This study showed that the neuromorphic adaptive control algorithm could learn over time and move between the target destination points faster compared to execution time for previous iterations. We showed that the robot arm learns to complete tasks 15\% faster in specific experiment scenarios such as scenarios with six or nine random target points (Table.~\ref{tab:freq}). This result indicates that there are combinations of random target points for which the algorithm learns to complete the task faster. In a follow on study, we plan to investigate how these particular scenarios generalize - specifically, what is it about a target selection strategy that makes it more or less suitable for accelerated learning and how to create pre-training strategies to select random target points that best support the learning capability of the algorithm.

\bibliographystyle{ACM-Reference-Format}
\bibliography{sample-authordraft}










\end{document}